\documentclass[11pt]{article}

 \usepackage[final]{acl}

\usepackage{times}
\usepackage{latexsym}

\usepackage[T1]{fontenc}

\usepackage[utf8]{inputenc}

\usepackage{microtype}

\usepackage{inconsolata}

\usepackage{graphicx}
\usepackage{amsmath}
\usepackage{amsfonts}
\usepackage{multirow}
\usepackage{booktabs}
\usepackage[table]{xcolor}
\title{Causal Evidence Extraction and Triangulation in Crisis Reports using Large Language Models: A ReliefWeb-based Study}

\author{Yuanjun Zhang \\
   University of Oulu, Finland \\
  LUT University, Finland\\
  \texttt{yuanjun.zhang@lut.fi} \\\And
   Mourad Oussalah\thanks{Corresponding author} \\
  University of Oulu, Finland \\
  \texttt{mourad.oussalah@oulu.fi} \\}

\begin{document}
\maketitle

\begin{abstract}
Humanitarian reports are long, noisy, and multi-topic, making it difficult to consolidate decision-relevant causal evidence. We present a ReliefWeb study (2000--2024) and a two-stage Large Language Model (LLM) pipeline that extracts structured intervention-outcome records with direction and strength attributes. Query-conditioned extraction restricts output to a specified intervention class, reducing retrieval-induced over-extraction, while snippet grounding links each relation to supporting text for auditability and classification. In an expert-annotated dataset of 100 reports, the best closed-source LLM achieved a weighted F1 score of 90.73\% with strong cost-efficiency, while Llama-3.1-8B with supervised fine-tuning reached 94.15\% weighted F1 score. We further propose context-preserving triangulation that aggregates strength-weighted evidence within disaster$\times$source cells, applies Laplace smoothing and equally weights cells to quantify cross-context convergence via a Level-of-Evidence score. Applied to cash assistance, food-related outcomes show strong positive convergence (LoE=0.865) and stable long-horizon trajectories. 
\end{abstract}

\section{Introduction}
\label{sec:Introduction}
Humanitarian crises generate large volumes of narrative situation reports describing interventions and evolving outcomes, yet decision-relevant evidence about \emph{what works, where, and under which constraints} remains hard to consolidate at scale.
Cash assistance illustrates the challenge: it is widely deployed, but reported impacts vary with market conditions, targeting, delivery mechanisms, and protection risks \cite{calp2015mpg,doocy2017cash,freccero2019safer,burton2020doing}.
A bottleneck is translating long, multi-topic narratives into structured, auditable intervention, and outcome evidence that can be compared across contexts.

Because keyword retrieval prioritizes recall, it often surfaces documents where the queried intervention is marginally mentioned. Besides, unconstrained extraction can then amplify this noise by producing irrelevant relations. Recent LLMs make prompt-based information extraction practical at scale \cite{brown2020gpt3}, and grounding conclusions in explicit textual evidence is increasingly emphasized for auditability \cite{deyoung2020eraser,lewis2020rag}.
However, evidence-synthesis extraction frameworks are typically developed for more topically clean input \cite{shi2025evidence}, motivating report-centric and question-aware designs.

\begin{figure*}[t]
  \centering
  \includegraphics[width=0.99\textwidth]{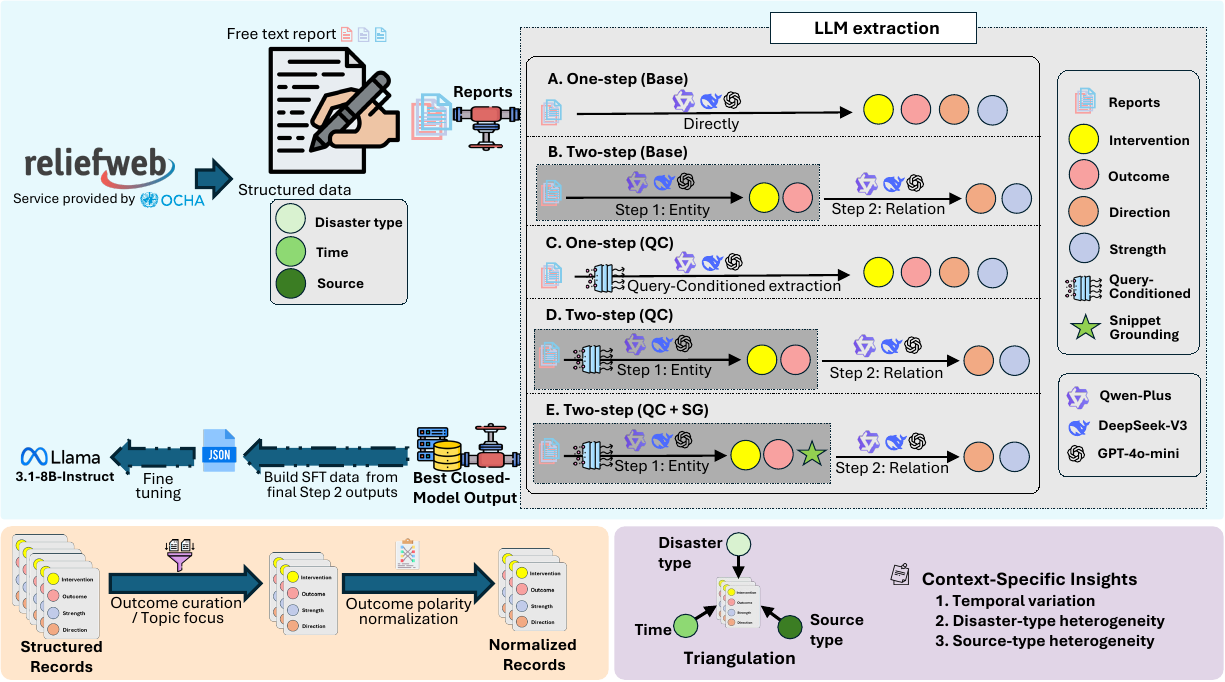}
    \caption{Overall framework for causal evidence extraction and context-preserving triangulation.
    ReliefWeb reports and metadata (disaster type, source type, time) are processed by a two-stage LLM pipeline with query conditioning and snippet grounding to produce structured causal records (Section~\ref{subsec:extraction}). We then curate outcomes and normalize polarity for consistent directionality, and aggregate evidence via cell-based triangulation over disaster$\times$source contexts using strength-weighted counts, Laplace smoothing, and equal cell weighting, yielding global probabilities ($P_P$, $P_Z$, $P_N$), a Level-of-Evidence (LoE) score, and temporal trajectories (Section~\ref{subsec:triangulation}).}

  \label{fig:overall-framework}
\end{figure*}

We address this gap with a two-stage LLM extraction pipeline for \emph{causal evidence extraction} from ReliefWeb reports \cite{reliefwebapi2025} in Fig.~\ref{fig:overall-framework}.
In \textbf{Stage~1}, \emph{query-conditioned extraction} restricts outputs to interventions in a query-defined class and extracts intervention--outcome candidates together with supporting snippets, reducing over-extraction from low-precision retrieval.
In \textbf{Stage~2}, \emph{snippet-grounded relation classification} predicts direction and strength for each candidate pair using the supporting snippets, improving auditability and inference.
For synthesis, we then introduce \textbf{context-preserving triangulation}, which aggregates strength-weighted directional evidence within disaster$\times$source cells and measures cross-context convergence rather than pooling raw counts \cite{jick1979triangulation}.
In addition, for lightweight deployment, we study an \textbf{optional distillation step}: using the best closed-source pipeline outputs as supervision, we apply Low-Rank Adaptation (LoRA) \cite{DBLP:conf/iclr/HuSWALWWC22} supervised fine-tuning to Llama-3.1-8B-Instruct \cite{grattafiori2024llama}.

Using ReliefWeb corpus (2000--2024), we compare five strategies across three API LLMs and an open-weight Llama-3.1-8B-Instruct model with LoRA fine-tuning.
Across settings, query conditioning yields large, consistent gains, and snippet grounding further improves faithfulness.
We also demonstrate long-horizon triangulation for cash assistance, producing interpretable global probabilities and temporal evidence trajectories.

\section{Related Work}
\label{sec:Related_Work}

Prior humanitarian natural language processing (NLP) research highlights domain shift and the need for trustworthy decision support \cite{rocca2023nlp_humanitarian}.
We study \emph{report-centric} extraction of intervention--outcome records from long, multi-topic ReliefWeb reports \cite{reliefwebapi2025}, leveraging document-level IE/RE \cite{yao2019docred} and evidence grounding \cite{deyoung2020eraser}.

Complementary to text-based evidence extraction, recent work has also emphasized reliability and explainability in disaster-related multimodal decision support, showing that preference optimization and explainable vision-language reasoning can improve both assessment performance and interpretability in disaster severity assessment \cite{ZHANG2026112674}. LLMs are increasingly used for evidence extraction and synthesis, including directional triangulation \cite{shi2025evidence} and end-to-end evidence-synthesis pipelines \cite{wang2025accelerating}.
We adapt this line to humanitarian reports with query-conditioned extraction and context-preserving synthesis within disaster$\times$source cells, while attaching short supporting spans for traceability. LLM factuality and hallucination detection have also been studied, e.g., via consistency-based checks \cite{manakul-etal-2023-selfcheckgpt}.

More specifically, our work differs from prior evidence extraction studies in three ways. First, we target long, noisy, multi-topic humanitarian reports. Second, we use query-conditioned extraction to control retrieval-induced over-extraction. Third, beyond extraction, we introduce context-preserving triangulation over disaster$\times$source cells, together with snippet grounding for auditability, instead of relying only on pooled document-level evidence.

\section{Methodology}
\label{sec:method}

Given a ReliefWeb report $x$ and a query $q$ specifying an intervention \emph{class}
(e.g., \textit{cash assistance}),
we extract a set of directional causal records
\begin{equation}
e = (a, o, \delta, \sigma, z)
\label{eq:record}
\end{equation}
where $a$ is an intervention mention (restricted to class $q$), $o$ is an outcome mention, $\delta\in\{\textsc{Inc},\textsc{Dec},\textsc{No}\}$ denotes whether the reported outcome increases, decreases, or shows no clear change, $\sigma\in\{\textsc{Weak},\textsc{Mod},\textsc{Strong}\}$ is the evidence strength, and $z$ is a supporting text snippet for auditability.

\paragraph{Data.} We query ReliefWeb with the keyword \textit{'cash assistance'} and retain English reports with accessible text (2000--2024).
We restricted to ReliefWeb-defined disaster and source types: four disasters (\textsc{Flood}, \textsc{Drought}, \textsc{Epidemic}, \textsc{Earthquake}) and five sources
(\textsc{Academic}, \textsc{Government}, \textsc{Media}, \textsc{NGO}, \textsc{Red Cross/Red Crescent Movement}), yielding 8,029 reports; \textsc{International Organization} and \textsc{Other} are excluded due to imbalance/ambiguity (Appendix~\ref{app:data_filtering} and Appendix~\ref{sec:AdditionalAnalyses}).
For evaluation, experts annotate 100 sampled reports (220 relations); see Appendix~\ref{app:annotate100} for details.

\paragraph{Evaluation.} We evaluate intervention/outcome extraction via semantic set matching using BERTScore (threshold 0.8)~\cite{zhang2019bertscore}.
Direction and strength are scored on matched intervention--outcome pairs; strength is scored only when direction is correct.
Since strength is ordinal (\textsc{Weak}$<$\textsc{Mod}$<$\textsc{Strong}), adjacent confusions receive partial credit (0.5).

\subsection{Query-conditioned and snippet-grounded extraction}
\label{subsec:extraction}

As shown in Fig.~\ref{fig:overall-framework}, our final method is a \textbf{two-stage extraction pipeline} with two LLM calls.
\textbf{Stage~1 (QC entity \& evidence extraction):} the LLM extracts only interventions in class $q$ (query-conditioned, QC), the associated \emph{outcome mentions} expressed in the report as raw text spans / free-text phrases, and supporting snippets $z$. The output of Stage~1 therefore consists of candidate intervention--outcome pairs together with their evidence snippets.
\textbf{Stage~2 (SG relation classification):} given these candidate pairs and snippets, the LLM predicts the relation labels, i.e., direction $\delta$ and strength $\sigma$. Snippets provide explicit grounding (snippet-grounded, SG) while the full report remains available.

Separately, we compare \textbf{single-prompt} versus \textbf{two-prompt} prompting as an ablation, and we also ablate QC and SG. Here, \emph{stage} refers to the two LLM calls in the final extraction pipeline, whereas \emph{single-prompt} and \emph{two-prompt} refer only to prompting variants.

Using the best \textbf{two-stage QC+SG extraction pipeline} as the teacher, we select the best-performing proprietary closed-source LLM.
We then decompose its outputs into \textbf{Stage-1 supervision} (QC intervention--outcome pairs with grounded snippets) and \textbf{Stage-2 supervision} (direction/strength conditioned on Stage~1), and train two parameter-efficient LoRA adapters (one per stage) on an open-weight Llama-3.1-8B-Instruct student while preserving the same two-stage inference structure.
To prevent leakage, we exclude the 100 evaluation reports from distillation; to reduce model-family relatedness bias, we choose a student outside the teacher's family \cite{wang-etal-2025-truth}.

\begin{table}[]
\centering
{\small
\begin{tabular}{l c c c c}
\toprule
\textbf{Model} & \textbf{Domain} & \textbf{GT} & \textbf{Base} & \textbf{QC} \\
\midrule
\multirow{2}{*}{\textbf{\shortstack{qwen\\plus}}}
 & Medical        & 3.09 & 5.00  & --   \\
 & Humanitarian   & 2.20 & 13.76 & 2.79 \\
\midrule
\multirow{2}{*}{\textbf{\shortstack{gpt-4o\\mini}}}
 & Medical        & 3.09 & 4.33 & --   \\
 & Humanitarian   & 2.20 & 7.43 & 2.19 \\
\midrule
\multirow{2}{*}{\textbf{\shortstack{deepseek\\v3}}}
 & Medical        & 3.09 & 3.70  & --   \\
 & Humanitarian   & 2.20 & 10.50 & 1.60 \\
\bottomrule
\end{tabular}}
\caption{Extraction volume comparison across domains and models. Values are average \#relations/document. QC is applied only for humanitarian reports.}
\label{tab:qc_effect}
\end{table}

\subsection{Context-preserving triangulation}
\label{subsec:triangulation}

We synthesize records across contexts while preventing high-volume contexts from dominating.

\paragraph{Polarity normalization.}
Outcomes may be positively framed (e.g., \textit{food security}) or negatively framed (e.g., \textit{food insecurity}).
We assign polarity $\pi(o)\in\{+1,-1\}$ and flip $\textsc{Inc}\leftrightarrow\textsc{Dec}$ when $\pi(o)=-1$ (leaving \textsc{No} unchanged),
so a \emph{positive} relation always denotes improvement.
(For the case study, polarity is labeled for frequent outcomes, e.g., frequency $\ge 3$. See Appendix~\ref{app:polarity-frequent-outcomes} for details.)

\begin{table*}[]
\centering
{
\setlength{\tabcolsep}{4.5pt}
\small
\begin{tabular}{lccccccc}
\toprule
\textbf{Model} & \textbf{Intervention} & \textbf{Outcome} & \textbf{Direction} & \textbf{Strength} &
\textbf{Weighted F1} & \textbf{Unit price (USD)} & \textbf{Efficiency} \\
\midrule

\multicolumn{8}{l}{\textbf{DeepSeek-V3}} \\
One-step (Base)            & 44.20 & 58.80 & 93.77 & 89.63 & 67.58 & 0.707 & 0.956 \\
Two-step (Base)            & 34.69 & 57.09 & 94.46 & 89.51 & 64.33 & 0.707 & 0.910 \\
One-step (QC)              & 75.00 & 74.65 & 95.85 & 91.38 & 82.34 & 0.707 & 1.164 \\
Two-step (QC)              & 76.47 & 82.12 & 97.19 & 92.20 & 85.46 & 0.707 & 1.208 \\
\rowcolor{gray!20}Two-step (QC + SG)         & 87.44 & 87.86 & 92.88 & 90.02 & 89.17 & 0.707 & 1.261 \\
\midrule

\multicolumn{8}{l}{\textbf{GPT-4o-mini}} \\
One-step (Base)            & 48.24 & 62.52 & 92.14 & 82.35 & 68.13 & 0.375 & 1.817 \\
Two-step (Base)            & 49.05 & 67.43 & 91.06 & 75.10 & 68.18 & 0.375 & 1.818 \\
One-step (QC)              & 73.28 & 77.33 & 94.08 & 84.91 & 80.98 & 0.375 & 2.159 \\
Two-step (QC)              & 84.49 & 86.34 & 94.29 & 81.08 & 86.32 & 0.375 & 2.302 \\
\rowcolor{gray!40}Two-step (QC + SG)         & 87.50 & 88.83 & 94.17 & 77.55 & 87.24 & 0.375 & 2.326 \\
\midrule

\multicolumn{8}{l}{\textbf{Qwen-Plus}} \\
One-step (Base)            & 34.96 & 49.86 & 92.11 & 89.97 & 61.86 & 0.198 & 3.124 \\
Two-step (Base)            & 33.77 & 52.95 & 88.09 & 89.01 & 61.44 & 0.198 & 3.103 \\
One-step (QC)              & 85.59 & 84.88 & 95.48 & 91.58 & 88.55 & 0.198 & 4.472 \\
Two-step (QC)              & 87.88 & 88.63 & 93.71 & 90.44 & 89.78 & 0.198 & 4.534 \\
\rowcolor{gray!60}Two-step (QC + SG)         & 87.31 & 91.99 & 93.65 & 91.05 & 90.73 & 0.198 & 4.582 \\
\midrule

\multicolumn{8}{l}{\textbf{Llama-3.1-8B-Instruct} using Two-step (QC + SG)} \\
No SFT              & 62.72 & 71.69 & 89.04 & 84.96 &  75.12 & -- & -- \\
\rowcolor{gray!80}LoRA SFT         & 94.06 & 95.68 &  94.64 & 91.50 & 94.15 &-- & -- \\
\midrule

\textbf{Weights}           & 0.30   & 0.30   & 0.20   & 0.20   & & & \\
\bottomrule
\end{tabular}}
\caption{Overall performance comparison across models and extraction strategies. All F1 scores are in \%. Weighted F1 uses weights 0.30/0.30/0.20/0.20. Base denotes the baseline extraction strategy from \citet{shi2025evidence}.}
\label{tab:overall_results}
\end{table*}


\paragraph{Cell-based evidence aggregation.}
Let $\mathcal{D}$ denote the set of disaster types and $\mathcal{S}$ denote the set of source types. Each record inherits disaster type $d\in\mathcal{D}$ and source type $s\in\mathcal{S}$.
We define triangulation cells $c=(d,s)\in\mathcal{D}\times\mathcal{S}$ and map normalized directions to $r\in\{P,Z,N\}$
(positive / no-change / negative).
We convert strength to an ordinal weight $w(\textsc{Weak}){=}0,\; w(\textsc{Mod}){=}1,\; w(\textsc{Strong}){=}2$ :
\begin{equation}
C_r^{(c)}=\sum_{i\in c} w(\sigma_i)\,\mathbb{I}(r_i=r).
\label{eq:cell_counts}
\end{equation}

\paragraph{Smoothed cell probabilities.}
To avoid degenerate estimates in sparse cells, we apply Laplace smoothing ($\alpha{=}0.1$):
\begin{equation}
P_r^{(c)}=\frac{C_r^{(c)}+\alpha}{\sum_{r'\in\{P,Z,N\}} C_{r'}^{(c)}+3\alpha}
\label{eq:cell_prob}
\end{equation}
Cells with $\sum_{r} C_r^{(c)}=0$ are discarded. 

\paragraph{Equal cell weighting and Level of Evidence.}
Let $\mathcal{C}$ denote the set of retained non-empty cells. We average cell probabilities equally:
\begin{equation}
P_r=\frac{1}{|\mathcal{C}|}\sum_{c\in\mathcal{C}} P_r^{(c)}
\label{eq:global_prob}
\end{equation}
Cross-context convergence is measured by a Level of Evidence score:
\begin{equation}
\mathrm{LoE}=\frac{\max(P_P,P_Z,P_N)-\frac{1}{3}}{1-\frac{1}{3}} \in [0,1].
\label{eq:loe}
\end{equation}
Here, $P_P$, $P_Z$, and $P_N$ denote the global cell-averaged probabilities of positive / no-change / negative relationships. By construction, $P_P$, $P_Z$, and $P_N$ are non-negative and sum to 1, so $\max(P_P,P_Z,P_N)\in[1/3,1]$. Therefore, $\mathrm{LoE}\in[0,1]$ and cannot be negative.

\paragraph{Temporal triangulation.}
For evidence trajectories, we recompute Eqs.~\eqref{eq:cell_counts}--\eqref{eq:loe} cumulatively by year $t$
using records with $\mathrm{year}_i\le t$, producing $(P_P(t),P_Z(t),P_N(t),\mathrm{LoE}(t))$.

\section{Experiments and Results}
\label{sec:results}
We evaluate on four LLMs: \textit{Qwen-Plus}~\cite{bai2023qwen}, \textit{GPT-4o-mini}~\cite{hurst2024gpt}, \textit{DeepSeek-V3}~\cite{liu2024deepseek}, and Llama-3.1-8B-Instruct.

\subsection{QC reduces over-extraction}
Keyword retrieval is recall-oriented and often surfaces reports where the queried intervention is only marginally mentioned; baseline prompting then over-extracts query-irrelevant relations.
Tab.~\ref{tab:qc_effect} shows severe over-extraction on humanitarian reports, while query-conditioned extraction (QC) brings the volume close to the expected ground-truth level across closed-source models.
Medical numbers are from \citet{shi2025evidence}.

\subsection{Extraction performance: ablations over QC, SG, and prompting variants}
Tab.~\ref{tab:overall_results} reports extraction performance on the expert-annotated benchmark.
Across models, QC provides the largest gains, while prompt decomposition and snippet grounding (SG) yield additional consistent improvements.
Among closed-source  models, Qwen-Plus with two-step QC+SG achieves the best weighted F1 (90.73\%) and is used for downstream triangulation.
Efficiency is computed as $\text{Efficiency} = (\text{Weighted F1} / 100) \,/\, \text{unit price}$ (prices in Appendix Tab.~\ref{tab:api_costs}).
Our pipeline also provides supervision for distillation: LoRA fine-tuning Llama-3.1-8B-Instruct improves weighted F1 from 75.12\% to 94.15\% (Appendix~\ref{app:lora}).

\begin{figure}[]
  \centering
  \includegraphics[width=\columnwidth]{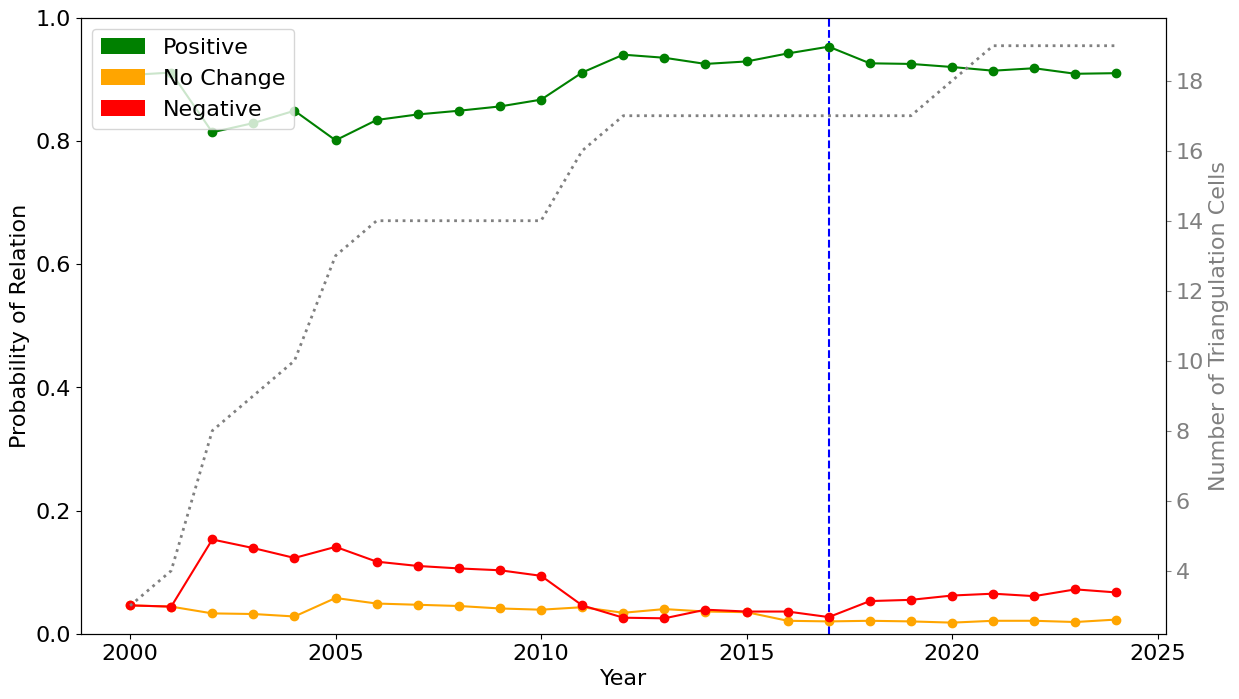}
  \caption{Cumulative temporal triangulation for polarity-normalized food-related outcomes:
cell-averaged $P_P$, $P_Z$, $P_N$ and the number of contributing disaster$\times$source cells over time.}
  \label{fig:triangulation_result}
\end{figure}

\subsection{Triangulation: cross-context convergence}
Using Qwen-Plus (two-step QC+SG), we extract 19{,}568 causal records across four disaster types (5{,}999 drought; 3{,}696 earthquake; 4{,}412 epidemic; 5{,}461 flood).
We focus on food-related outcomes due to high evidence density (Appendix~\ref{sec:AdditionalAnalyses}).
After polarity normalization (87 positive vs.\ 65 negative outcome strings with frequency $\ge3$), triangulation over $4\times5=20$ disaster$\times$source cells retains 19 non-empty cells and yields $P_P=0.91$ and $\mathrm{LoE}=0.865$.

Fig.~\ref{fig:triangulation_result} shows cumulative trajectories.
Early years (2000--2005) have sparse coverage and volatile estimates; from 2006--2017, expanding cell coverage yields strong positive convergence, peaking in 2017 ($P_P=0.953$, $\mathrm{LoE}=0.929$).
From 2018--2024, the signal remains highly positive with mild softening, consistent with residual context heterogeneity and operational ``do no harm'' constraints \cite{calp2015mpg,freccero2019safer,burton2020doing}.

\section{Conclusion}
We propose a query-conditioned, snippet-grounded two-stage extraction pipeline to extract auditable intervention--outcome evidence from long humanitarian reports, and a context-preserving cell-based triangulation method for cross-context synthesis.
On ReliefWeb, the best closed-source setting reaches 90.73\% weighted F1, LoRA SFT on Llama-3.1-8B-Instruct reaches 94.15\%, and triangulation shows strong positive convergence of cash assistance on food outcomes (LoE=0.865).

\section*{Limitations}

Due to limitations in funding and manpower, our corpus and case study are constrained by a keyword query (“cash assistance”), English-only accessible text, a limited set of disaster types, and a coarse source taxonomy, which may reduce generalizability to other interventions, languages, hazards, and reporting contexts. The expert-annotated evaluation set is also small; while it allows us to capture overall trends, performance estimates may be sensitive to sampling and annotation guidelines. Our triangulation and aggregation reflect convergence of reported evidence and rely on design choices (e.g., polarity normalization, weighting, smoothing) that can influence the resulting scores. We note that the strength labels in this work reflect the strength of reported evidence rather than effect magnitude or population-scale impact, which we leave for future work. Therefore, the outputs should be treated as decision support with human review rather than as definitive conclusions.

\section*{Ethical Considerations}
This work analyzes publicly available humanitarian reports and does not intentionally collect or infer sensitive personal information. Nevertheless, reports may contain incidental mentions of individuals or vulnerable groups; we therefore use the data only for research purposes and recommend caution when releasing derived artifacts. Because ReliefWeb serves as an upstream source, our method may inherit biases in the reporting ecosystem (e.g., uneven geographic coverage, organizational incentives, and language accessibility), which can affect extracted evidence and aggregated scores. In addition, assessments of humanitarian measures in text may themselves be subjective or contested, and LLM-based extraction may further reflect hidden or unknown model biases. To mitigate this, we attach supporting snippets for auditability and triangulate across disaster$\times$source contexts to reduce reliance on any single report or context. Therefore, the outputs should not be used as a stand-alone basis for operational choices. We position the system as an assistive tool to surface and summarize reported evidence, and we recommend human review—especially for contested or ambiguous claims and for any downstream decisions that may impact affected populations.

\section*{Acknowledgements}
This work is funded by the European Union's Horizon 2020 research and innovation programme under grant agreement No. 101004509, and the Collaboration of Humanities and Social Sciences in Europe (CHANSE) research project Digital Emergency Communication (DIGeMERGE).

\bibliography{custom}

\clearpage 
\appendix
\section{Data Filtering and Preprocessing}
\label{app:data_filtering}

This section describes the dataset before filtering, including the full set of reports retrieved from ReliefWeb,
the distribution across disaster types and source categories, and the filtering criteria applied to obtain the final
dataset used in our experiments.

\paragraph{Retrieval overview and disaster coverage.}
We first retrieve ReliefWeb reports using the keyword \textit{``cash assistance''} and keep English reports with accessible plain text (2000--2024).
Tab.~\ref{tab:disaster_distribution} summarizes the disaster-type distribution \emph{before} applying our final selection criteria.
The corpus spans a wide range of hazards, with \textsc{Flood}, \textsc{Epidemic}, \textsc{Drought}, and \textsc{Earthquake} being the most frequently covered types.
To ensure sufficient evidence density and stable cross-context analysis, we focus on these four high-coverage disaster types in subsequent experiments.

\begin{figure*}[]
  \centering
  \includegraphics[width=0.95\textwidth]{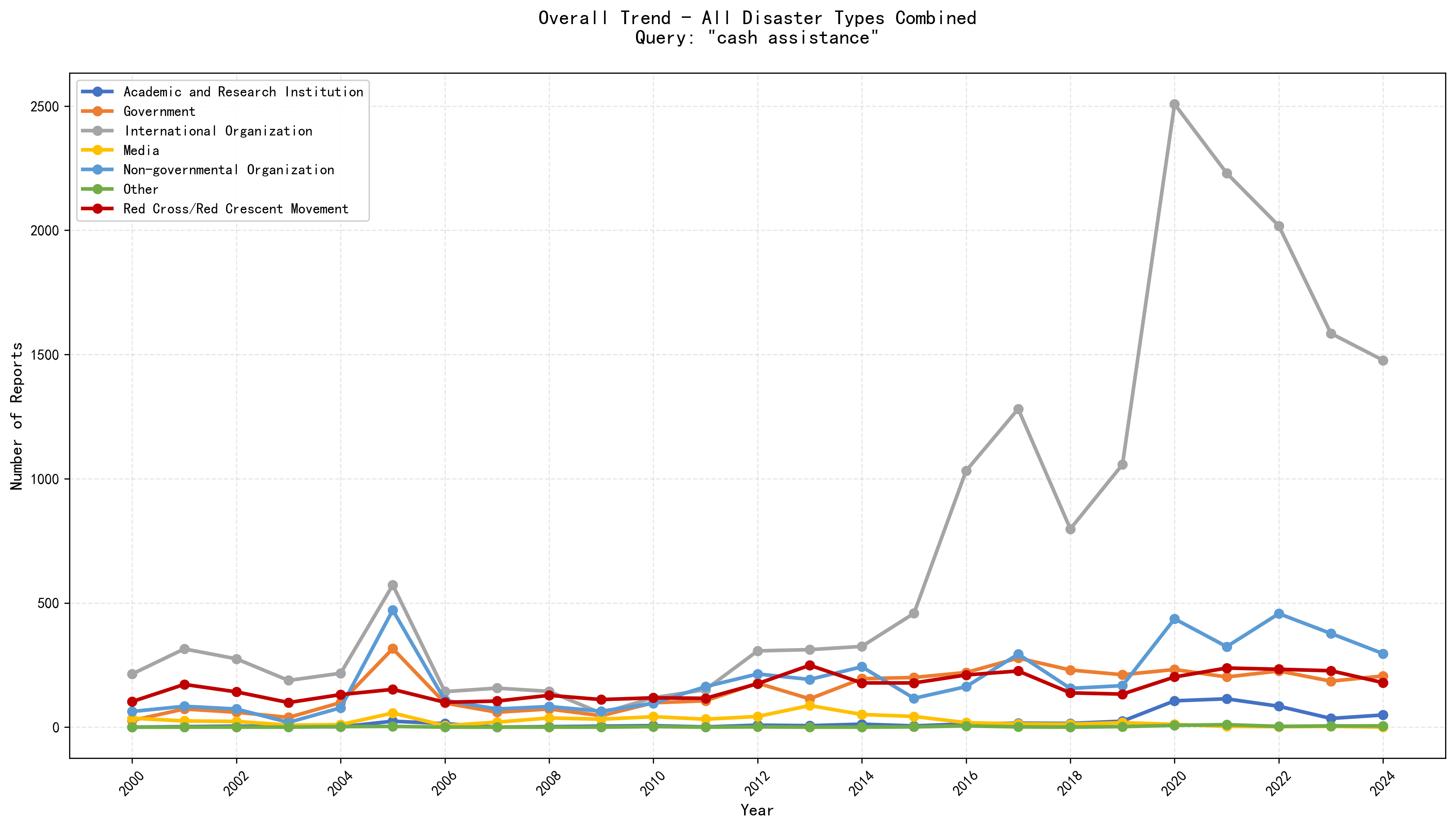}
  \caption{Annual number of ReliefWeb reports by source type.}
  \label{fig:overall_trend}
\end{figure*}

\paragraph{Source-type composition over time.}
Because reporting volume and institutional participation vary substantially across years, we further examine the annual distribution by source type.
Tab.~\ref{tab:source_year_distribution} provides pre-filter year-by-year counts for the source categories, highlighting both long-term growth in reporting and shifts in which institutions contribute reports.
Fig.~\ref{fig:overall_trend} visualizes the same trend.

\paragraph{Filtering rationale.}
Based on the above distributions, we apply filtering and selection to balance coverage and interpretability:
(i) retain only reports with accessible text and valid metadata,
(ii) restrict to the four most prevalent disaster types for stable cell-level analysis,
and (iii) map ReliefWeb sources into the source-type taxonomy used in the main experiments.
The resulting dataset forms the basis for both extraction evaluation and downstream triangulation.

\begin{table}[t]
\centering
\begin{tabular}{lr}
\toprule
\textbf{Disaster Type} & \textbf{Total Reports}  \\
\midrule
Flood & 5907  \\
Epidemic & 5878  \\
Drought & 5431  \\
Earthquake & 3108  \\
Tropical Cyclone & 2858  \\
Flash Flood & 2134 \\
Land Slide & 1860  \\
Other & 1322 \\
Tsunami & 895  \\
Insect Infestation & 511  \\
Severe Local Storm & 323  \\
Cold Wave & 319  \\
Volcano & 272  \\
Storm Surge & 265  \\
Technological Disaster & 263  \\
Mud Slide & 201 \\
Snow Avalanche & 79  \\
Wild Fire & 73  \\
Fire & 35 \\
Heat Wave & 34  \\
Extratropical Cyclone & 20  \\
\bottomrule
\end{tabular}
\caption{Distribution (before applying the final selection criteria) of ReliefWeb reports by disaster type (2000--2024).}
\label{tab:disaster_distribution}
\end{table}

\begin{table*}[]
\centering
\begin{tabular}{rccccccc}
\toprule
\textbf{Year} & \textbf{Academic} & \textbf{Government} & \textbf{Intl. Org.} & \textbf{Media} & \textbf{NGO} & \textbf{Other} & \textbf{RC/RC} \\
\midrule
2000 & 0 & 27 & 214 & 36 & 63 & 0 & 103 \\
2001 & 2 & 73 & 315 & 25 & 84 & 0 & 172 \\
2002 & 4 & 60 & 275 & 23 & 73 & 0 & 142 \\
2003 & 3 & 39 & 188 & 8 & 19 & 0 & 99 \\
2004 & 1 & 99 & 217 & 10 & 78 & 2 & 131 \\
2005 & 24 & 316 & 572 & 57 & 471 & 3 & 152 \\
2006 & 14 & 97 & 143 & 6 & 108 & 0 & 100 \\
2007 & 0 & 60 & 157 & 20 & 73 & 0 & 105 \\
2008 & 2 & 73 & 144 & 37 & 83 & 0 & 128 \\
2009 & 4 & 46 & 56 & 32 & 64 & 0 & 111 \\
2010 & 6 & 98 & 118 & 42 & 96 & 2 & 118 \\
2011 & 1 & 106 & 151 & 32 & 163 & 0 & 116 \\
2012 & 8 & 177 & 307 & 43 & 214 & 1 & 175 \\
2013 & 6 & 114 & 312 & 87 & 192 & 0 & 249 \\
2014 & 12 & 195 & 325 & 51 & 243 & 0 & 178 \\
2015 & 5 & 200 & 458 & 43 & 116 & 1 & 178 \\
2016 & 13 & 220 & 1032 & 18 & 163 & 5 & 210 \\
2017 & 16 & 278 & 1281 & 13 & 294 & 1 & 226 \\
2018 & 15 & 230 & 798 & 12 & 156 & 0 & 138 \\
2019 & 24 & 211 & 1057 & 18 & 167 & 2 & 133 \\
2020 & 106 & 232 & 2508 & 11 & 436 & 7 & 202 \\
2021 & 114 & 202 & 2230 & 3 & 324 & 10 & 238 \\
2022 & 84 & 226 & 2018 & 2 & 457 & 3 & 233 \\
2023 & 35 & 185 & 1585 & 3 & 377 & 5 & 227 \\
2024 & 49 & 206 & 1477 & 0 & 296 & 5 & 179 \\
\midrule
\textbf{Total} & 548 & 3770 & 17938 & 632 & 4810 & 47 & 4043 \\
\bottomrule
\end{tabular}
\caption{Annual number of ReliefWeb reports by source type (2000--2024). Academic = Academic and Research Institution; 
Government = Government; 
Intl. Org. = International Organization; 
Media = Media; 
NGO = Non-governmental Organization; 
Other = Other; 
RC/RC = Red Cross/Red Crescent Movement.}
\label{tab:source_year_distribution}
\end{table*}

\section{Additional Analyses}
\label{sec:AdditionalAnalyses}
\subsection{Outcome distribution}
\begin{figure*}[]
  \centering
  \includegraphics[width=0.7\textwidth]{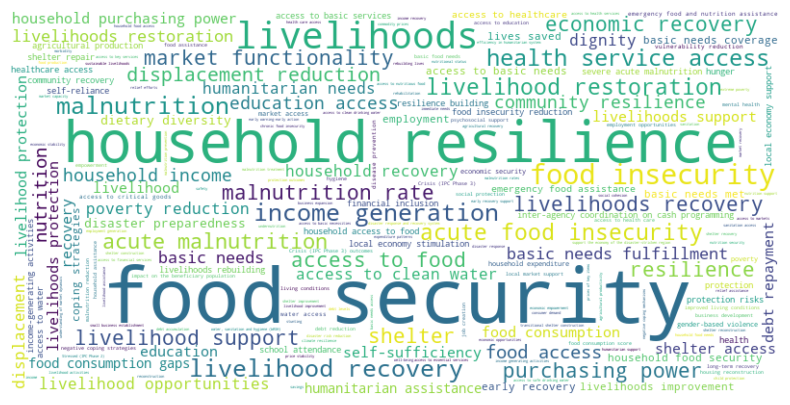}
  \caption{Frequency-based word cloud of extracted outcomes (used to motivate the food-related case study).}
  \label{fig:outcome_cloud}
\end{figure*}

\paragraph{Motivation for the food-related case study.}
To choose an outcome family with sufficient evidence for long-horizon triangulation, we inspect the frequency distribution of extracted outcome mentions.
Fig.~\ref{fig:outcome_cloud} shows a frequency-based word cloud of outcomes extracted by our best-performing configuration, revealing that food-related outcomes appear prominently.
This observation motivates the focus on food-related outcomes in the main triangulation analysis (Section~\ref{subsec:triangulation}).

\subsection{API pricing for efficiency metric}
\paragraph{Cost accounting.}
To compare practical deployability across LLMs, we report a cost-efficiency metric in Tab.~\ref{tab:overall_results} based on per-token API prices.
Tab.~\ref{tab:api_costs} lists the input/output prices used in our calculations (per 1M tokens), and we convert qwen-plus and deepseek-v3 pricing from RMB using a fixed exchange rate stated in the caption.

\begin{table*}[]
\centering
\begin{tabular}{lccc}
\toprule
\textbf{Model} & \textbf{Input cost (USD)} & \textbf{Output cost (USD)} & \textbf{Unit price (USD, avg)} \\
\midrule
gpt-4o-mini & 0.15  & 0.60  & 0.375 \\
qwen-plus   & 0.113 & 0.283 & 0.198 \\
deepseek-v3 & 0.283 & 1.131 & 0.707 \\
\bottomrule
\end{tabular}
\caption{API pricing used in cost-efficiency calculations (prices per 1M tokens). And qwen-plus/deepseek-v3 are converted from RMB using 1 RMB = 0.1414 USD.}
\label{tab:api_costs}
\end{table*}

\section{Sampling Details of the Expert Evaluation Set}
\label{app:annotate100}

\paragraph{Sampling procedure.}
The expert evaluation set consists of 100 full-length crisis reports randomly sampled from our 8029-report corpus. We use simple random sampling at the report level, without stratification by source type, year, or region. Because each report is typically long and may contain multiple interventions and outcomes, expert annotation in this setting is substantially more time- and expertise-intensive than standard single-label annotation.

\paragraph{Source-type composition of the sampled reports.}
As a descriptive check, the 100 sampled reports cover multiple major source categories in the corpus, including Red Cross/Red Crescent Movement (34 reports), Non-governmental Organization (32 reports), Government (25 reports), Media (5 reports), and Academic and Research Institution (4 reports). Since the evaluation set is obtained through simple random sampling rather than stratified sampling, we do not expect its source-type proportions to exactly match the corpus-level distribution. Instead, this breakdown is reported to show that the sampled reports span the major source categories represented in the corpus.

\section{Polarity Labeling for Frequent Outcomes}
\label{app:polarity-frequent-outcomes}

\paragraph{Rationale for labeling polarity only for frequent outcomes.}
In the triangulation case study, we label polarity only for outcome terms occurring at least 3 times. This decision is motivated by the frequency distribution of outcome terms. Specifically, outcome terms with frequency $\geq 3$ comprise only 941 unique vocabularies, yet they cover 13554 out of 19568 samples (69.27\%). By contrast, outcome terms occurring only 1--2 times include 5080 unique vocabularies, but together they cover only 6014 out of 19568 samples (30.73\%). This distribution reflects a typical long-tail pattern: a relatively small set of frequent outcomes covers the majority of the samples, whereas a large number of rare outcomes contributes limited sample coverage. Given our effort under limited manpower, we prioritize the high-frequency outcomes in polarity labeling. We further clarify that these polarity labels are used only in the triangulation case study and do not affect the main extraction evaluation.

\section{Supplementary Discussion and Targeted Future Directions}

Our results show that query-conditioned extraction and snippet-grounded relation classification can make long humanitarian reports more auditable and more amenable to context-preserving triangulation. Several broader methodological directions also suggest how this framework could be strengthened in future work.

Retrieval reliability and clue selection remain key bottlenecks for long-document humanitarian evidence extraction. Our current corpus construction relies on keyword-based retrieval, which is practical but can still surface reports in which the queried intervention is only marginally discussed. More broadly, retrieval-augmented modeling has long motivated modular and interpretable use of external evidence \cite{guu2020realm}. For our setting, future versions of the pipeline could incorporate retrieval-quality control and contextual filtering before extraction. Context-driven index trimming may help remove query-inconsistent retrieved contexts, while compact clue selection can reduce reasoning cost by retaining only the minimum sufficient evidence \cite{ma-etal-2024-context,10.1145/3774904.3792158}. Query clustering and hybrid search further suggest richer retrieval over both free text and structured metadata such as source type, disaster type, and time \cite{liu2025queries,STABLE}. Adaptive retrieval and self-reflective evidence use are also relevant: rather than always applying the same fixed pipeline depth, a future system could decide when extra retrieval or verification is necessary and remain robust when some retrieved content is irrelevant \cite{asai2024selfrag,zhang2026expseekselftriggeredexperienceseeking,yoran2024robust}.

Ensuring faithful, focused, and robust evidence grounding is equally important in this setting. Our current method attaches supporting snippets and performs relation classification with access to both snippets and the full report, but future systems could evaluate more explicitly whether intermediate reasoning remains faithful to the cited evidence. Work on faithfulness-aware planning and acting is directly relevant here because it emphasizes evidence-constrained reasoning rather than relying only on final-answer quality \cite{li2025faithact}. Related work on hallucination analysis in retrieval-augmented systems also suggests the value of finer-grained checks for unsupported or weakly supported outputs \cite{niu-etal-2024-ragtruth}. In parallel, robust composed retrieval studies highlight transferable ideas for evidence calibration and focus control, such as aligning heterogeneous signals and reducing the influence of noisy or peripheral content \cite{ReTrack,OFFSET}. Related work on noise mitigation and differential semantics is also relevant when multiple snippets or contexts must be combined without collapsing distinct signals into a single oversimplified summary \cite{INTENT,REFINE}.

Further improvements are also likely to come from a more data-centric treatment of both the extraction model and the distillation pipeline. Recent work shows that gains can come not only from larger models, but also from better supervision design, stronger filtering, and more robust treatment of noisy data. For our setting, this suggests several practical extensions: synthesizing or augmenting training instances for rare intervention--outcome patterns, curating higher-quality teacher data, and selecting fine-tuning examples based on difficulty and utility \cite{liu2025synthvlmhighqualityefficientsynthesis,lin2026mmfinereasonclosingmultimodalreasoning,ge2026emsdialogsyntheticmultipersonemergency}. Retrieval-based noisy-label detection provides a natural starting point for identifying suspicious labels or low-quality teacher supervision \cite{liu2023retrieval}. Progressive robust learning under noisy correspondence further suggests that student training may benefit from improved filtering, curriculum design, or confidence-aware weighting when teacher outputs are imperfect \cite{HABIT}.

As retrieval-enhanced evidence systems mature, questions of provenance, release control, modular orchestration, and stable model characterization become increasingly important \cite{ma2026stableexplainablepersonalitytrait}. Methods for detecting unauthorized use of retrieval-enhanced data suggest that future evidence platforms may need stronger provenance protection for curated corpora, extracted records, and derived knowledge stores \cite{liu2025stole}. More broadly, high-stakes multi-agent work suggests a natural architectural extension of our current multi-stage pipeline: separate agents could specialize in retrieval, verification, grounding, normalization, and triangulation, while a higher-level controller dynamically allocates trust, requests more evidence, or routes difficult cases for human review \cite{long2026emomasemotionawaremultiagenthighstakes,long2025emodebtbayesianoptimizedemotionalintelligence}. In our context, such an architecture may be more appropriate than a single monolithic extractor for handling long reports, heterogeneous evidence, and uncertainty-sensitive humanitarian decision support.

\section{LoRA Fine-Tuning Details}
\label{app:lora}

\paragraph{Implementation details (LoRA SFT).}
We fine-tune an open-weight \textit{Llama-3.1-8B-Instruct} with LoRA \cite{DBLP:conf/iclr/HuSWALWWC22} and train two adapters to match our two-step pipeline (for Step~1 and Step~2). For both adapters, we use rank $r{=}8$ (\texttt{lora\_target=all}), cutoff length 4096, 3 epochs, learning rate $1\mathrm{e}{-4}$.

Training runs on a single node with 3$\times$V100 GPUs; we evaluate/save every 111 steps and load the best checkpoint by dev loss. Distillation data are built from a disjoint training split and exclude the 100 expert-annotated evaluation reports.

\section{Summary of Notations}
\label{app:notations}

For readability, Tab.~\ref{tab:notation_summary} summarizes the main notations used in this paper.

\begin{table*}[t]
\centering
\begin{tabular}{p{0.30\textwidth} p{0.62\textwidth}}
\toprule
\textbf{Notation} & \textbf{Meaning} \\
\midrule
$q$ & Query specifying the intervention class (e.g., \textit{cash assistance}). \\
$e$ & A structured causal record extracted from a report. \\
$(a,o,\delta,\sigma,z)$ & The tuple form of a causal record defined in Eq.~\eqref{eq:record}. \\
$a$ & Intervention mention, restricted to the query-defined class $q$. \\
$o$ & Outcome mention extracted from the report. \\
$\delta \in \{\textsc{Inc}, \textsc{Dec}, \textsc{No}\}$ & Direction label: increase, decrease, or no clear change. \\
$\sigma \in \{\textsc{Weak}, \textsc{Mod}, \textsc{Strong}\}$ & Strength label of the reported evidence. \\
$z$ & Supporting text snippet grounding the extracted relation. \\
$\pi(o) \in \{+1,-1\}$ & Polarity of outcome $o$ used for direction normalization. \\
$\mathcal{D}$ & Set of disaster types. \\
$\mathcal{S}$ & Set of source types. \\
$d \in \mathcal{D}$ & A disaster type associated with a record. \\
$s \in \mathcal{S}$ & A source type associated with a record. \\
$c=(d,s)$ & A triangulation cell defined by a disaster--source pair. \\
$\mathcal{C}$ & Set of retained non-empty triangulation cells. \\
$r \in \{P,Z,N\}$ & Normalized relation category: positive, no-change, or negative. \\
$P$ & Positive relation after polarity normalization. \\
$Z$ & No-change relation after polarity normalization. \\
$N$ & Negative relation after polarity normalization. \\
$w(\sigma)$ & Ordinal weight assigned to strength label $\sigma$. \\
$w(\textsc{Weak})=0$ & Weight for weak evidence. \\
$w(\textsc{Mod})=1$ & Weight for moderate evidence. \\
$w(\textsc{Strong})=2$ & Weight for strong evidence. \\
$C_r^{(c)}$ & Strength-weighted count of relation type $r$ in cell $c$ in Eq.~\eqref{eq:cell_counts}. \\
$\mathbb{I}(\cdot)$ & Indicator function. \\
$P_r^{(c)}$ & Smoothed probability of relation type $r$ within cell $c$ in Eq.~\eqref{eq:cell_prob}. \\
$\alpha$ & Laplace smoothing parameter; in this paper, $\alpha = 0.1$. \\
$P_r$ & Global cell-averaged probability of relation type $r$ in Eq.~\eqref{eq:global_prob}. \\
$P_P, P_Z, P_N$ & Global probabilities of positive, no-change, and negative relations. \\
$t$ & Year index used in cumulative temporal triangulation. \\
\bottomrule
\end{tabular}
\caption{Summary of the main notations used in the paper.}
\label{tab:notation_summary}
\end{table*}

\end{document}